\pdfoutput=1

\documentclass{article}
\usepackage[margin=1in]{geometry}
\usepackage{url}
\usepackage{graphicx}
\usepackage{amsmath}
\usepackage{amssymb}
\usepackage{setspace}
\usepackage{hyperref}
\usepackage{subcaption}

\usepackage{float} 
\usepackage{placeins} 

\title{\textbf{Southern Newswires: A Large-Scale Study of Mid-Century Wire Content Beyond the Front Page}}
\author{\textbf{Michael McRae} \\
Trinity College Dublin\\
\texttt{mcraemi@tcd.ie}
}
\date{}

\begin{document}
\maketitle

\begin{abstract}
This paper describes the construction of a large-scale corpus of historical wire articles from U.S. Southern newspapers, spanning 1960–1975 and covering multiple wire services (e.g., Associated Press, United Press International, Newspaper Enterprise Association). Unlike prior work that focuses primarily on front-page content, the corpus captures wire-sourced articles across the entire newspaper, offering broader insight into mid-century Southern news coverage. The analysis incorporates both raw OCR text and a version processed through an LLM-based text correction pipeline designed to reduce OCR noise and improve suitability for quantitative text analysis. Multiple versions of the same wire dispatch are retained, allowing for the study of editorial differences in language and framing across newspapers. Articles are classified by wire service, enabling comparative analysis of editorial patterns across agencies. Together, these features provide a detailed perspective on how Southern newspapers transmitted national and international news during a transformative period in American history.
\end{abstract}

\vspace{1em}

\section{Introduction}
From the mid-20th century onward, U.S. newspapers relied extensively on wire services---especially the Associated Press (AP), United Press International (UPI), and the Newspaper Enterprise Association (NEA)---to provide timely national and international coverage to local communities. In Southern states, where local editors balanced syndication with the idiosyncrasies of regional politics and culture, wire content both shaped and reflected how readers engaged with pivotal events such as the Civil Rights Movement, Cold War politics, and the Vietnam War. By distributing stories to numerous subscribing papers, these agencies effectively curated much of the region's daily reading material, from front-page headlines to interior briefs and columns. 

Existing digitisation efforts have made strides in extracting and analysing historical newspaper content, but they often concentrate on front-page articles or single wire services, leaving potentially significant material in the interior pages underexplored. Moreover, while large corpora of historical text now exist, many remain burdened by noisy optical character recognition (OCR), complicating fine-grained content analysis. Further, previous approaches often discard or merge multiple reprints of the same wire text into one observation, neglecting the observed phenomena that local editors often abridged or otherwise edited wire text while retaining the agency byline, generating subtle nuances in linguistic choice. 

In this paper, I present the construction and analysis of the \texttt{Southern Newswire Corpus}, a large-scale research corpus, a novel dataset that moves beyond these limitations in four ways: 1) it captures wire articles throughout the newspaper (not just the front page), vastly increasing the scope of wire-sourced material; 2) it includes a \emph{raw} OCR version and a second \emph{cleaned} version, generated by a large language model (LLM) configured for text correction, substantially reducing OCR-induced noise; 3) it explicitly identifies major wire agencies (AP, UPI, NEA), facilitating cross-agency comparisons of coverage, tone, and editorial policy; and 4) reprints of the same wire text are treated uniquely. Each article is assigned an \textit{article\_nw\_id} that denotes which wire dispatch it ultimately derives from. 

I focus on Southern newspapers from 1960 to 1975, years marked by pivotal events such as the Civil Rights Movement, Vietnam War escalation, and significant shifts in local and national politics. Although these newspapers were published in the Southern United States, the wire articles they carried came from national and international bureaus, providing coverage of a wide range of global issues. This combination offers a particularly rich vantage point for understanding how wire content—originating from distant locations or national headquarters—was communicated to local communities and shaped the public discourse in the South during a time of profound social and political change.


Section~\ref{sec:dataset} overviews the resulting dataset structure and key statistics; Section ~\ref{sec:methodology} detail methods for layout detection, OCR, duplication identification, and service tagging; and Section~\ref{sec:discussion} discusses applications, limitations, and ethical considerations.

\section{Related Work}
\label{sec:related}

Large-scale digitisation efforts such as Chronicling America \cite{chroniclingamer} have made vast collections of historical U.S. newspaper scans available, prompting the development of methods to transform raw page images into structured text corpora. Early pipeline designs emphasized layout segmentation and OCR; for example, Tesseract has been widely used for text extraction from scanned pages, while more specialized Transformer-based OCR approaches \cite{dell2024americanstories} have recently emerged to tackle noisier archives. These processes typically produce article-level content by detecting bounding boxes for headlines and paragraphs.

Once article texts are extracted, another important challenge is reproduced-text detection, essential for identifying syndicated material like wire stories. Earlier approaches, such as the Viral Texts project \cite{smith2015}, relied on n-gram overlaps to uncover reprinted content within historical collections. More recent methods leverage neural encoders to map near-duplicate articles into similar vector representations, improving robustness to noise or minor edits. Contrastive training of bi-encoder models has proven especially effective \cite{sentencebert}, as illustrated by Dell and colleagues’ wire-article clustering pipeline. Furthermore, large language models (LLMs) now offer a potent means of post-OCR text cleanup, reducing character-level error rates and enhancing downstream tasks like entity recognition \cite{brown2020language}. Combined, these advances in layout segmentation, deduplication, and correction underpin the construction of high-quality historical text datasets, including wire-focused resources such as \texttt{Headlines} \cite{headlines2024}, \texttt{Newswire} \cite{dell2024newswire}, and the present \texttt{Southern Newswire Corpus}, which applies these methods to an expanded set of newspapers and services while adding a fully corrected text option.

\section{Dataset}
\label{sec:dataset}
\subsection{Source Newspapers and Date Range}
The newswire articles in \texttt{Southern Newswire Corpus} are constructed by detecting articles written by newswire services AP, UPI and NEA, in approximately 10 million digitised newspaper pages from Southern U.S. publications \footnote{In this context, the South represents the Confederate States of America e.g., South Carolina, Mississippi, Florida, Alabama, Georgia, Louisiana, Texas, Virginia, Arkansas, Tennessee, and North Carolina.} between 1960 and 1975. The newspapers include daily print newspapers as well as smaller weekly printed newspapers. All pages of each edition available are included in the pipeline. Each article is available in a \textit{raw} OCR version as well as an \textit{LLM-corrected} version, which reduces OCR noise to facilitate quantitative text analysis. The dataset covers multiple wire services, primarily AP, UPI, and NEA. Summary of the sample are pr0vided in \autoref{tab:dataset_counts}.

\begin{table}[H]
\centering
\begin{tabular}{l r}
\hline
\textbf{Metric} & \textbf{Count} \\
\hline
Total articles & 57,547,723 \\ 
Newswire articles & 9,571,254 \\ 
AP articles &  7,429,704\\ 
UPI articles &  2,024,254\\ 
NEA articles &  117,296\\ 
Unique newswire articles & 1,768,567 \\ 
Unique AP articles &  1,326,036\\ 
Unique UPI articles & 393,015 \\ 
Unique NEA articles & 49,516 \\ 
\hline
\end{tabular}
\caption{Summary of article counts in the dataset.}
\label{tab:dataset_counts}
\end{table}

\begin{figure}[H]
\centering
\includegraphics[width=0.6\textwidth]{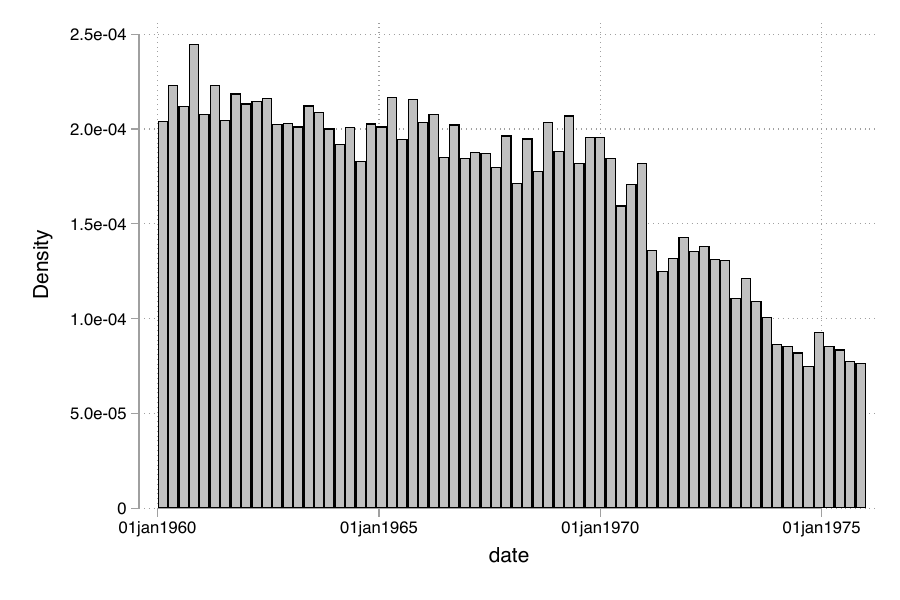}
\caption{Density of all articles by date}
\label{fig:articles_density}
\end{figure}

\begin{figure}[htbp]
  \centering
  \begin{minipage}[b]{0.4\textwidth}
    \centering
    \subfloat[Newswire vs. All Articles Over Time]{%
      \includegraphics[width=\textwidth]{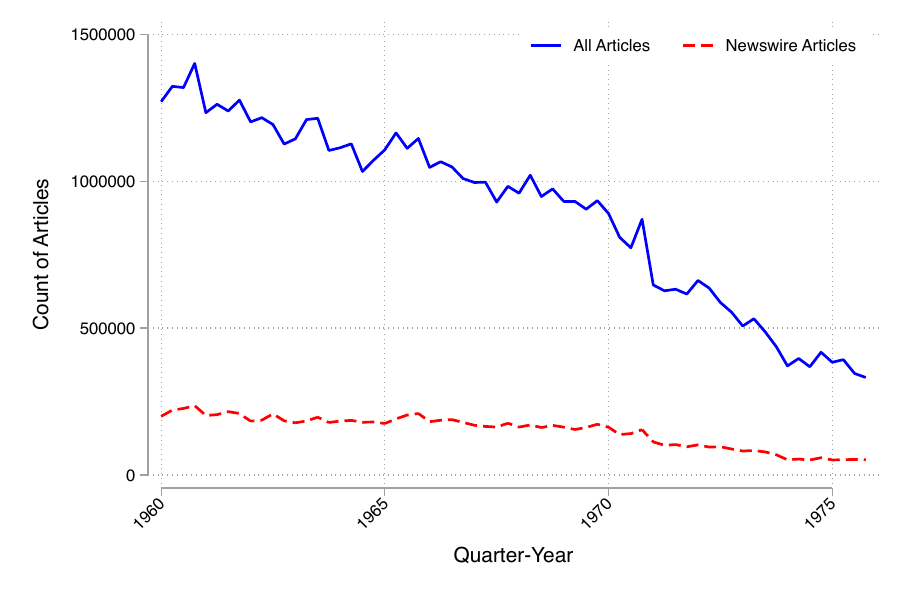}%
      \label{fig:nw_all_1}
    }
  \end{minipage}
  \hspace{2em}
  \begin{minipage}[b]{0.4\textwidth}
    \centering
    \subfloat[AP, UPI, and NEA Proportions Over Time]{%
      \includegraphics[width=\textwidth]{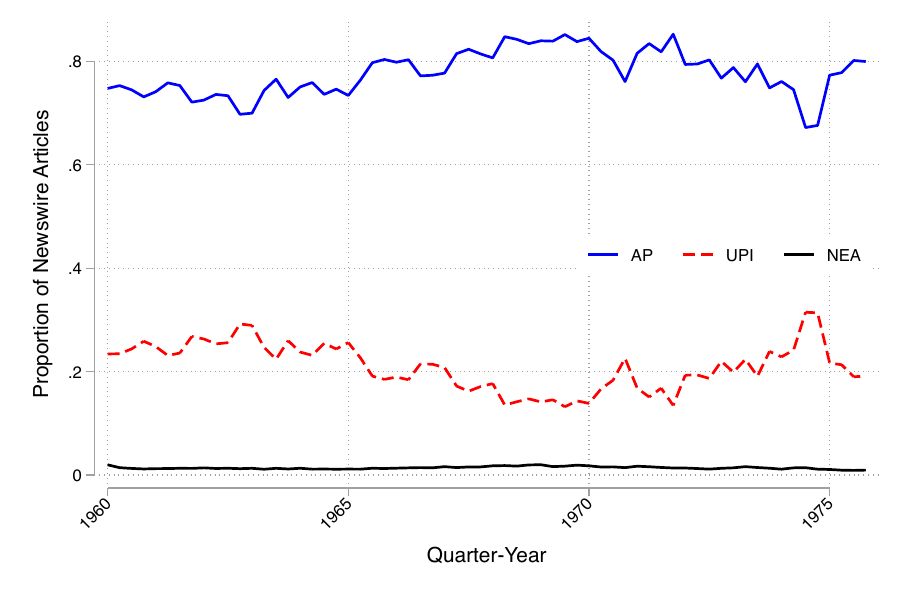}%
      \label{fig:nw_all_2}
    }
  \end{minipage}
  \caption{Evolution of newswires (1960–1975)}
  \label{fig:nw_evolutions}  
\end{figure}

\begin{figure}[htbp]
\centering
\includegraphics[width=0.6\textwidth]{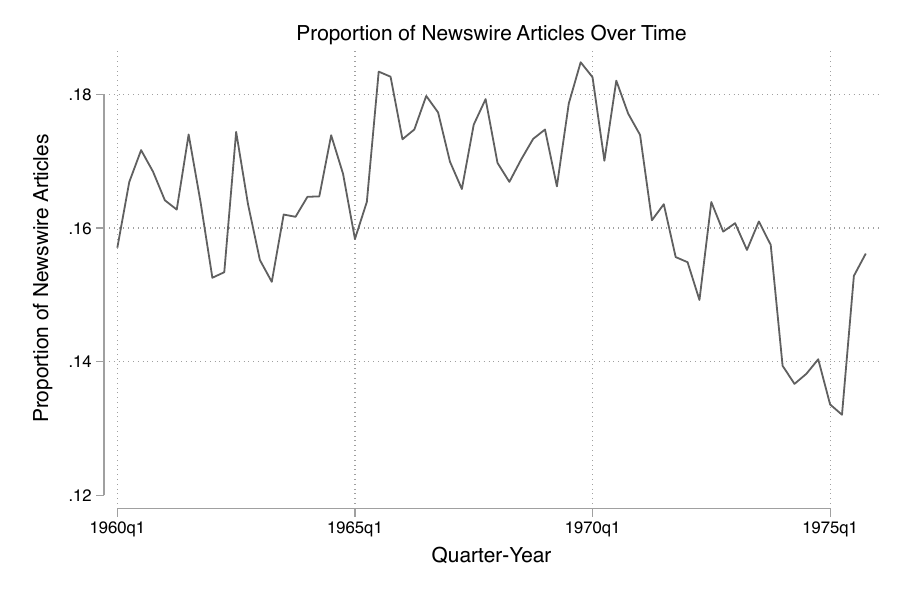}
\caption{Evolution of proportional newswires out of all articles (1960 - 1975).}
\label{fig:nw_proportions}
\end{figure}

\autoref{fig:articles_density}  shows that the overall article density is higher in the mid-1960s before tapering toward the early 1970s, while \autoref{fig:nw_all_1} indicates that the total number of articles in Southern newspapers declines from around 1.5 million in 1960 to roughly half that amount by the mid-1970s, with the count of wire articles following a similar (but lower) trajectory. \autoref{fig:nw_all_2} then breaks down wire articles by agency, revealing AP’s steady dominance relative to UPI and NEA over the 15-year span. In addition, the proportion of total articles that are wire-sourced, hovers between 15\% and 18\% in the early 1960s, briefly climbing above 18\% around the mid-decade, as shown in \autoref{fig:nw_proportions}. After 1965, however, the proportion trends downward, dipping to near 12\% in the early 1970s before partially rebounding by 1975. This decline suggests that while wire content remained an important component of Southern newspapers throughout the period, its relative share of total articles in \texttt{Southern Newswire Corpus} gradually fell over time.

\begin{figure}[htp]
\centering
\includegraphics[width=0.7\textwidth]{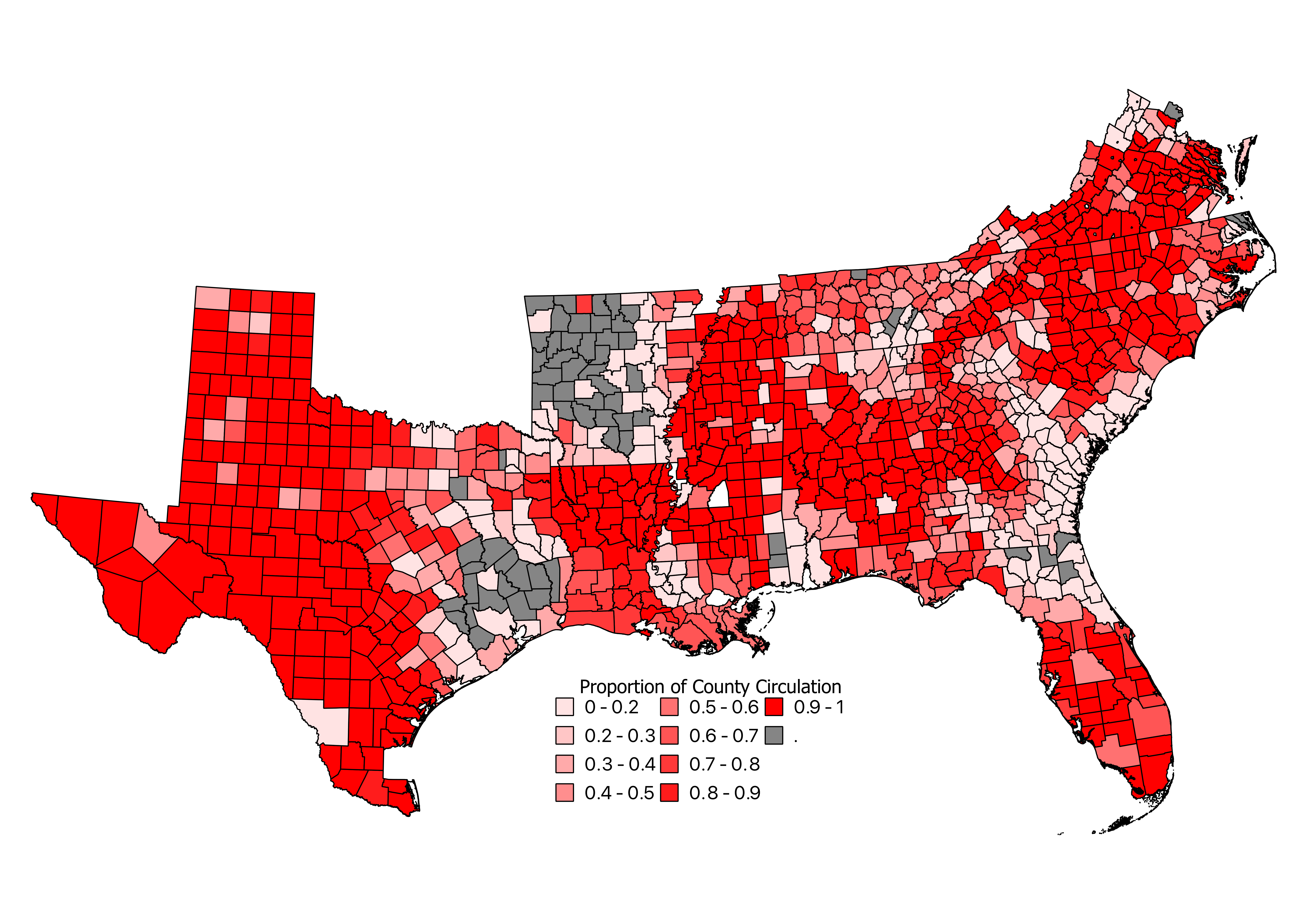}
\caption{\small{Proportion of total county daily newspaper circulation in 1964 by newspapers in \texttt{Southern Newswire Corpus}. Calculated from novel dataset of daily newspaper circulation
collected by the Audit Bureau of Circulation (ABC) from 1964}}
\label{fig:county_circulation}
\end{figure}

In addition to article text and newswire metadata, I provide the newspaper source of each version of each article, including geo-referencing of newsroom headquarters. Utilising a novel dataset of daily newspaper circulation collected by the Audit Bureau of Circulation (ABC) from 1964, I calculate that the newspapers in \texttt{Southern Newswire Corpus} account for more than 50\% of total daily circulation for 70\% of all Southern counties (see \autoref{fig:county_circulation}). 

To generate topics for newswire articles, I employ a Latent Dirichlet Allocation (LDA) model to label article topic area for each article in \texttt{Southern Newswire Corpus}. I conduct a multi-sampling process whereby I divided the corpus into multiple sub-samples and trained an LDA model on each, extracting topic distributions from every sub-sample. Results are then aggregated across sub-samples for a unified topic representation and weighted based on topic prevalence. The most representative 50 topics from the combined distributions are selected and OpenAI’s GPT-4o model is utilised to assign human-readable labels to these topics. To simplify topics, I grouped closely related topics into broader categories. I experimented with transformer-based embeddings using BERTopic. Despite the method's strength in capturing semantic relationships, BERTopic produced a high level of noise, with many clusters lacking clear thematic boundaries and exhibiting significant overlap. The increased granularity, while informative in some contexts, resulted in less coherent and less stable topic assignments compared to other methods. 

\begin{figure}[htp]
\centering
\includegraphics[width=0.8\textwidth]{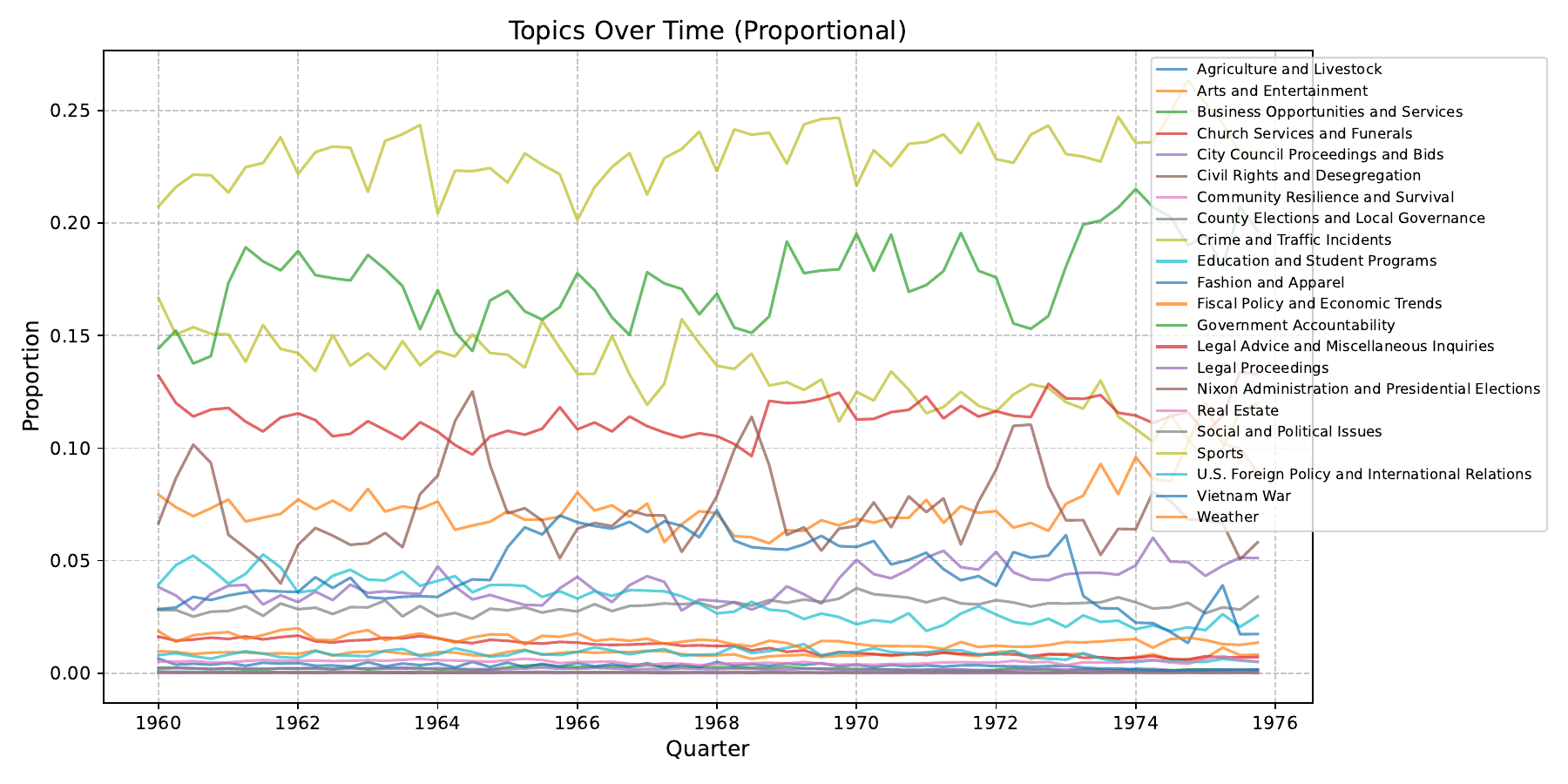}
\caption{\small{Proportion of topics over time of newswire articles}}
\label{fig:topics}
\end{figure}

\section{Methodology}
\label{sec:methodology}

\subsection{Layout Analysis and OCR}
To detect articles, headlines, advertisements, and other content regions in newspaper scans, I employ a layout detector trained on YOLOv10 (Medium). I train the layout detection model on a set of 4,000 labeled newspaper images containing 88974 annotated layout objects (e.g., articles, headlines, images, advertisements, masthead). Legibility classification is then conducted using an existing image classifier with a Mobilenet v3 backbone from the \texttt{American Stories} project \cite{dell2024americanstories}. After filtering out illegible blocks, recognised text regions undergo OCR via Tesseract. Finally, if an article spans multiple bounding boxes (e.g., a headline and an adjacent text block), I apply a rule-based association that uses bounding-box coordinates to merge them into a single structured observation, preserving the headline, author byline (where identifiable), and article text.

\subsection{Identifying Wire Services}
Each recognised article is classified by wire service (AP, UPI, or NEA) using three lightweight Sentence-BERT-based models, each fine-tuned to detect a single service from its textual cues. For training data, I collected articles with explicit bylines referencing one of the three agencies and supplemented them with unlabelled articles that mention typical dateline or style markers (e.g., “(AP)”, “(UPI)”, “From Our NEA Staff”), followed by manual verification. Training data includes 4000 articles: 1000 for each service and 1000 locally produced articles. Each model encodes the article text (including the headline and any byline) into a Sentence-BERT embedding \cite{reimers2019}, and a simple linear classifier predicts whether that article matches the wire service in question. Articles flagged by more than one model or not exceeding a minimum confidence threshold in any model are excluded from the dataset, minimising the possibility of incorrect wire-service attribution. In practice, this conservative filtering strategy removes roughly 4\% of candidate wire articles. The remaining items are assigned a wire service label and thus included in the final \texttt{Southern Newswire Corpus}.

\subsection{Identifying Shared Wire Dispatches}
To facilitate analysis of reprinted wire stories without discarding local modifications, I adopt the noise-robust de-duplication approach of Silcock et al.~\cite{silcock2023noise}. Their model, initially fine-tuned on historical newspaper data, encodes each article's text into a Sentence-BERT-like embedding. I then use approximate nearest-neighbour retrieval to locate articles with high semantic similarity published around the same date. Unlike a traditional deduplication pipeline, which might unify all reprints into a single “canonical” text, this method assigns a shared \textit{article\_nw\_id} to each cluster. This preserves every local version of a wire story, enabling comparison between local versions of the same underlying dispatch.

\subsection{LLM-Based Text Correction}
To provide a more refined version of the dataset, I developed a text-correction pipeline that uses Llama\,3.2, a large language model configured for minimal rewriting. Specifically:
\begin{itemize}
    \item \textbf{Error Reduction.} Llama\,3.2 corrects common OCR artefacts such as misread characters, broken words, or stray punctuation, improving text clarity without introducing anachronistic spellings.
    \item \textbf{Format Preservation.} I preserve paragraph structure and avoid altering genuine historical language. Corrections are limited to unambiguous OCR errors, ensuring that archaic or dialect terms remain intact.
\end{itemize}
In practice, this LLM-based approach significantly boosts entity recognition accuracy and overall readability, while retaining the historical integrity of the text.
%

\section{Discussion and Future Directions}
\label{sec:discussion}
\subsection{Research Applications.} 
By expanding beyond the front page, this dataset captures a broader range of historical themes that may not have been headline news but still circulated regionally. The corrected version supports robust NLP tasks such as topic modelling, network analysis of named entities, and large-scale language modelling fine-tuning on mid-century content. Cross-agency tagging lets researchers compare editorial or rhetorical variations among AP, UPI, and NEA.

\subsection{Limitations.} 
1) Coverage Gaps: Some newspapers or date ranges remain undigitised. 2) OCR Artefacts: While LLM corrections significantly help, some subtle errors persist. 3) Historical Biases: Language from this era can include biases or insensitive terminology, which I have deliberately not filtered to maintain authenticity.

\subsection{Future Work.} 
Moving forward, this dataset will be broadened to all US states from 1960 - 1975. I also intend to refine the text correction approach using open-source models, allowing entirely local processing of historical text rather than reliance on proprietary LLMs.

\section{Conclusion}
By capturing wire-sourced material across entire newspapers, distinguishing among multiple agencies, offering both raw and LLM-corrected versions, and maintaining variations of underlying wire dispatches, the \textbf{Southern Newswire Corpus} expands the opportunities for historical text analysis. Scholars from computational linguistics, digital humanities, and social sciences can now more easily explore how wire news was transmitted and transformed in the American South during a period of profound social and political change. This work demonstrates how large-scale historical wire corpora can be constructed and analysed to support replication-oriented research.

\section*{Acknowledgments}
I gratefully acknowledge prior research on large-scale wire detection and historical newspaper digitisation, which inspired and informed the methods used here. I also thank institutions that support large-scale historical text digitisation.. 

\bibliographystyle{abbrv}

\section{appendix}
\subsection{Code Availability}
Code used for the analyses presented in this paper is available from the author upon reasonable request.

\begin{center}
\url{https://mike-mcrae.github.io/}
\end{center}


\subsection{Newswire Classifiers}

I train and release three separate BERT-based classifiers to identify whether a newspaper article was produced by one of three major newswire services: \textit{Associated Press (AP)}, \textit{United Press International (UPI)}, and \textit{Newspaper Enterprise Association (NEA)}. Each model is a fine-tuned version of \texttt{bert-base-uncased} from the Hugging Face Transformers library, designed for binary classification with labels indicating the presence (1) or absence (0) of a specific newswire source. 

The classifiers were trained on historical newspaper articles from the period 1960--1975 using a labeled corpus of public-domain articles. Each model was trained on 4,000 articles per round (1,000 from the target newswire and 3,000 from other sources). To enhance performance on short news snippets, we trained using only headlines, bylines, and the first 100 characters of articles. The training was conducted using a TPU (v2-8) in Google Colab over four epochs with a batch size of 64 and a learning rate of $2 \times 10^{-5}$, optimized with AdamW and a binary cross-entropy loss function.

The models achieve high classification performance, with F1-scores of 0.9925 for the AP model, 0.9999 for UPI, and 0.9876 for NEA, making them suitable for large-scale classification tasks involving historical news archives. The classifiers are publicly available on Hugging Face and can be accessed at:

\begin{center}
\url{https://huggingface.co/mikemcrae25/newswire_classifiers}
\end{center}

\end{document}